# The Perceptron Algorithm: Image and Signal Decomposition, Compression, and Analysis by Iterative Gaussian Blurring

by

**Vassilios S. Vassiliadis, Ph.D.**

*Senior Lecturer*

**Department of Chemical Engineering,
University of Cambridge,
Pembroke Street,
Cambridge CB2 3RA,
United Kingdom**

**13 January 2006**

## Abstract

A novel algorithm for tunable compression to within the precision of reproduction targets, or storage, is proposed. The new algorithm is termed the "Perceptron Algorithm", which utilises simple existing concepts in a novel way, has multiple immediate commercial application aspects as well as it opens up a multitude of fronts in computational science and technology. The aims of this paper are to present the concepts underlying the algorithm, observations by its application to some example cases, and the identification of a multitude of potential areas of applications such as: image compression by orders of magnitude, signal compression including sound as well, image analysis in a multilayered detailed analysis, pattern recognition and matching and rapid database searching (e.g. face recognition), motion analysis, biomedical applications e.g. in MRI and CAT scan image analysis and compression, as well as hints on the link of these ideas to the way how biological memory might work leading to new points of view in neural computation. Commercial applications of immediate interest are the compression of images at the source (e.g. photographic equipment, scanners, satellite imaging systems), DVD film compression, pay-per-view downloads acceleration and many others identified in the present paper at its conclusion and future work section.

## 1. Introduction: image compression and the Perceptron Algorithm

Image compression nowadays is a crucial aspect in many commercial and scientific applications, where the primary aim is the reduction of size of an image file so as to minimize the storage and transmission requirements. Most algorithms of commercial value and standard operate at a bitwise manner, e.g. JPEG, TIFF with various components for compression, etc.

The point of view adopted in this work is to look at an image "as an image" and apply principles related to how humans perceive a field of view. For example, mammalian reactions when presented with a rapidly changing scene are to perceive in high speed situations as "friend of foe", and to react accordingly (albeit erroneously at times). Where more detail is studied in a field, it appears that biological processing requires more time to react. For example, in cases where one individual can afford this, more time is spent studying a scene to resolve and process finer detail. This is contrasted with high speed changes in the field which are perceived much faster, as mentioned, although motion blurring restricts the perception of finer detail. In other words for faster processing of changing scenes speed of reaction is paramount and for this a blurred image seems to suffice.

Motivated by this, we conducted some experimental steps of a new algorithm which is described mathematically in the following simple schema:

$$
\begin{aligned}
&\text{Blur}_{(i)} = \text{Gaussian\_Blurring}(\text{Image}_{(i)}, \text{noise}_{(i)}, \text{blur\_coefficient}_{(i)}); \\
&\text{Compressed\_Blur}_{(i)} = \text{Compression}(\text{Blur}_{(i)}); \\
&\text{Image}_{(i+1)} = \text{Image}_{(i)}(-)\text{Compressed\_Blur}_{(i)} \\
&i = 1, 2, \ldots, N
\end{aligned}
\tag{1}
$$

The algorithm in equations (1) in our implementation uses the following details, implemented in GIMP version 2.0 produced by the GNU free software organisation:

- "Gaussian_Blurring(.,.,.)" is the standard Gaussian blurring function provided in GIMP.
- "noise" is a small amount of "Spread" added to the images prior blurring which is used as a heuristic technique to reduce artefacts introduced by high JPEG compression (see below). If the noise of the current working image is already perceived as "high" then no further noise is added to the image prior to blurring.
- "blur_coefficient" is the standard deviation in pixels used in the application of Gaussian blurring. This initially is set to something rather high, approximately used as ½ of the maximum dimension of the image (vertical or horizontal) in pixels, and it is reduced by a reduction factor greater than 1 from one iteration to the next, usually we use a factor of 2, for example building a sequence of blur coefficients in the form of {1000, 500, 250, 125, 60, 30, 15, 8, 4, 2, 1}, and usually stopping at a blur coefficient of 1.
- "Compression(.)" is a standard function used from the options in GIMP. Although the original image and all working images $\text{Image}_{(i)}$ are used in "raw" format (TIFF uncompressed is used in our examples), the stored blurred

- images and the ones subtracted from the working image are low grade JPEG of a few Kbytes in size.
- The subtraction operation, "(-)", in the outlined algorithm above uses any suitable "subtraction" operation to take away the compressed low grade blurred image from the working image, resulting in a new working image. In GIMP for (-) we use the "Grain Extract" operation in the "Layers" superposition mode selection.

Upon completion, the operation of the algorithm results in *N* low grade compressed images, $\text{Compressed\_Blur}_{(i)}$, and a high grade working image, $\text{Image}_{(N)}$, which serves as a base image. The reconstruction without any loss whatsoever of the original image can be constructed by the following equation:

$$\boxed{\text{Image}_{(1)} = \sum_{i=1}^{N} \text{Compressed\_Blur}_{(i)} (+) \text{Image}_{(N+1)}} \tag{2}$$

Where the (+) is some appropriate addition of the images in equation (2), reversing the operation (-). The same operation is implied in the summation symbol. In GIMP we have used the reverse of "Grain Extract" which is "Grain Merge".

Now, since the above equation reverses the entire process, the original image is reconstructed flawlessly. The compressed blur images can be as small as desired, using the JPEG operation compression ratio/quality to derive an acceptable size file for the "layer" generated. Each compressed blurred image will be referred to from here on as a layer, while their collection along with the final base image will be referred to as the Gaussian Blurring decomposition stack. The problem arises with the base image which is still in uncompressed form, and any lossy compression on it will reduce the quality of the reconstructed image. It is noted that if one is working in raw format (e.g. TIFF) the base image will be of the same size as the original image and hence storing it as is would defy the purpose of compression; for analysis purposes though (see later discussion) this might be desired.

To control this aspect of loss in reconstruction, two remedies are possible at least (when significant overall compression is the aim):

- Increase the number of *N* compressed blurred layers, by reducing the factor of blurring coefficient, so as to produce more closely packed compressed blurred images, and
- Compress the final base image but with a higher quality compression so as to reduce artefacts introduced in it, if JPEG is used for it as well for example.

## 2. General signal compression with the "Perceptron Algorithm"

Quite clearly, the ideas above can be used to compress any kind of signal data series. For example, images can be viewed as two-dimensional collections of information, while for example audio files can be viewed as one-dimensional collections of information with x-coefficient the time axis. A similar schema to equations (1) is outlined for CDaudio files (generic term used here) and the application for example of MP3 compression for the compression of the *blurred* signal. Gaussian blurring, as diffusion of datum point values, can be applied in the same fashion as in equations (1) to the audio data series:

$$\begin{aligned}
&\text{Blur}_{(i)} = \text{Gaussian\_Blurring}(\text{CDaudio}_{(i)}, \text{noise}_{(i)}, \text{blur\_coefficient}_{(i)}); \\
&\text{Compressed\_Blur}_{(i)} = \text{Compression}(\text{Blur}_{(i)}); \\
&\text{CDaudio}_{(i+1)} = \text{CDaudio}_{(i)} (-) \text{Compressed\_Blur}_{(i)} \\
&i = 1, 2, \ldots, N
\end{aligned} \quad (1)$$

For Compressed_Blur(.) function we could use here MP3 low-sampling compression. The reconstruction of the signal would require the simultaneous un-compression of the low grade MP3 files in the stack, along with that of the base file as well, so as to reconstruct point-by-point the original signal values. Of course, again the issue of appropriate compression of the base file arises, but this may be controlled in the same way as suggested in the discussion for image reconstruction above. The remainder of the paper focuses exclusively on image compression and analysis. Future work will focus also on the general properties of Gaussian Blurring and analysis for general type signals.

## 3. Example case studies

Two case studies are presented in this section highlighting the observed properties when the algorithm is applied to real world images. The first case study shows the decomposition stack and the resulting reconstruction of the image in two ways: bottom up, and top down, for an original image which is noise-free at large. The second application works on a noisy original image, which shows how the algorithm seems to accumulate all noisy elements in the second half (bottom half) of the stack, lending itself potentially as an ideal noise reduction scheme where noise is reduced predominantly in the layers where it is accumulated.

**3.1 Decomposition of a noise-free image**

This image was produced in 16-bit colour depth TIFF from CANON EOS20D, by converting it from its RAW format (the CANON RAW is 12-bit colour depth). The TIFF was mapped to 8-bit colour depth as use of JPEG compressions restricts colour depth to 8-bit.

The first shown below in Table 1 is the decomposition of the original image into the Perceptron Algorithm stack.

**Table 1: Decomposition of a noise-free image**

| | |
|---|---|
| Blur 1:<br><br>Parameters in pixel values:<br>spread = 30, diffusion = 250<br>file size = 72.2 Kbytes<br><br>It is noted that the first blurred layer of the Perceptron Algorithm effectively removes all the colour from the image, while all remaining images in the stack become increasingly greyer. | 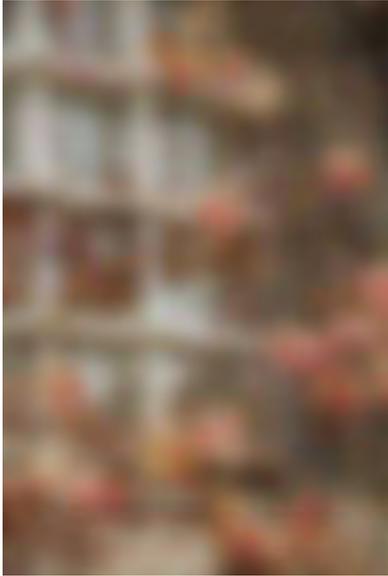 |
| Blur 2:<br><br>Parameters in pixel values:<br>spread = 30, diffusion = 250<br>file size = 65.5 Kbytes | 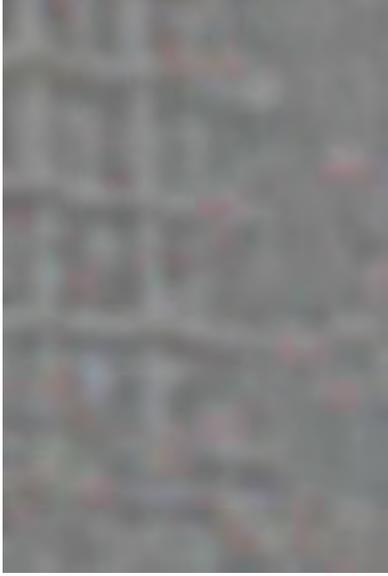 |
| Blur 3:<br><br>Parameters in pixel values:<br>spread = 30, diffusion = 250<br>file size = 65.7 Kbytes | 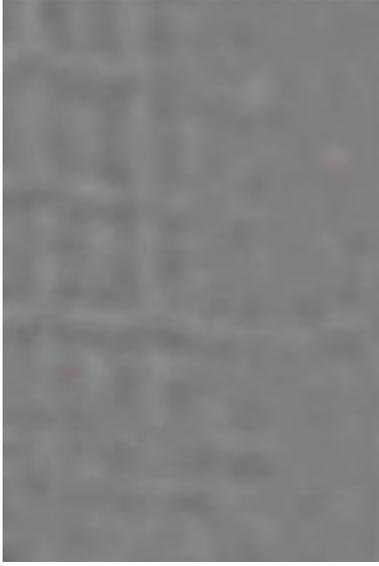 |
| Blur 4:<br><br>Parameters in pixel values:<br>spread = 30, diffusion = 150<br>file size = 73.6 Kbytes<br><br>This blur layer begins to show the thick edges detail, e.g. the bars in the window and the thick branches of the plant. This continues for layers Blur 5 and Blur 6. | 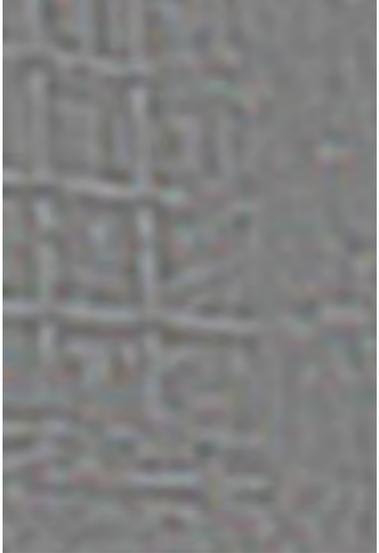 |
| Blur 5:<br><br>Parameters in pixel values:<br>spread = 30, diffusion = 100<br>file size = 89.6 Kbytes | 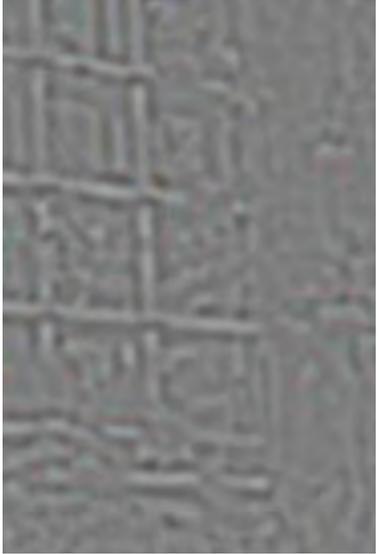 |
| Blur 6:<br><br>Parameters in pixel values:<br>spread = 30, diffusion = 70<br>file size = 86.5 Kbytes | 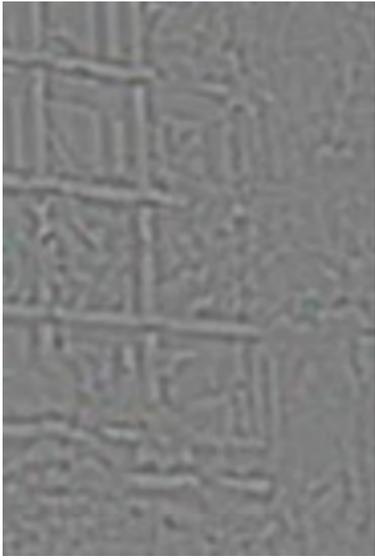 |
| Blur 7:<br><br>Parameters in pixel values:<br>spread = 30, diffusion = 50<br>file size = 84.1 Kbytes<br><br>The edges shown here begin to indicate finer detail both around the bars of the window as well as the outline of the leaves of the tree and finer branches. | 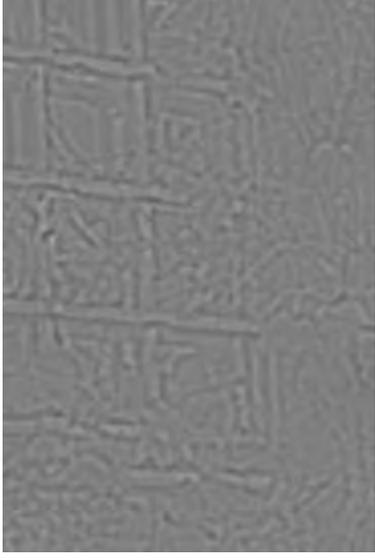 |
| Blur 8:<br><br>Parameters in pixel values:<br>spread = 30, diffusion = 30<br>file size = 115.0 Kbytes<br><br>This level of the decomposition along with Blur 9 and Blur 10 begin to focus on finer edge details. | 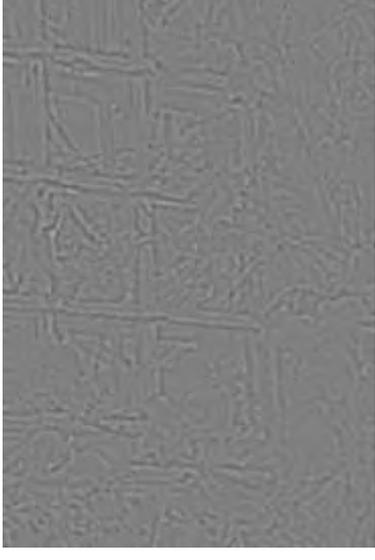 |
| Blur 9:<br><br>Parameters in pixel values:<br>spread = 20, diffusion = 20<br>file size = 116.0 Kbytes | 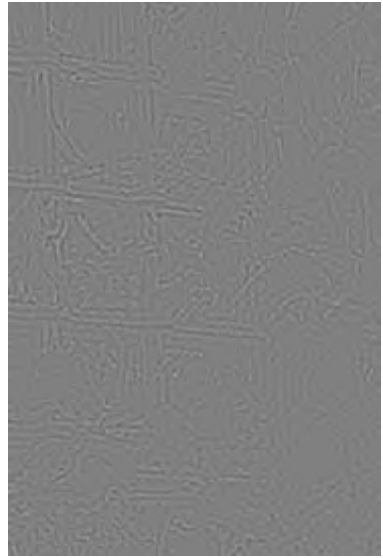 |
| Blur 10:<br><br>Parameters in pixel values:<br>spread = 10, diffusion = 15<br>file size = 308.0 Kbytes<br><br>This blur layer is dominated by very fine detail edges. Below the image is copied and enhanced in contrast to highlight this feature.<br><br>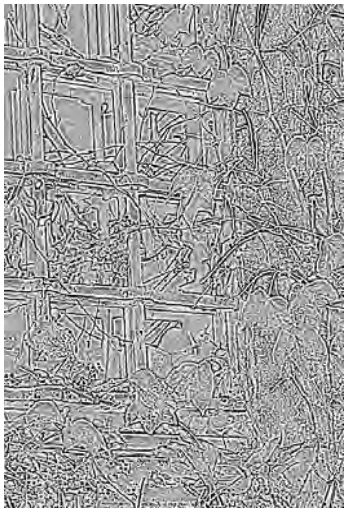 | 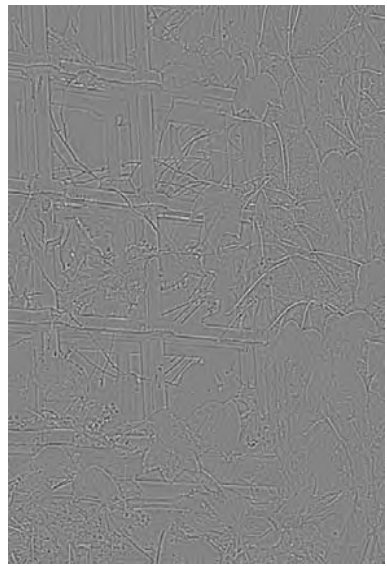 |
| Blur 11:<br><br>Parameters in pixel values:<br>spread = 5, diffusion = 7<br>file size = 174.0 Kbytes<br><br>This blur layer shows the edges beginning to show halos as the information diffuses into the grey background. | 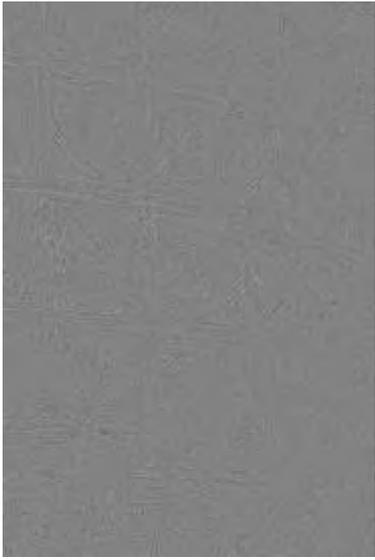 |
| Blur 12:<br><br>Parameters in pixel values:<br>spread = 3, diffusion = 5<br>file size = 152.0 Kbytes<br><br>From this level of blurring, and up to Blur level 13 and the Base image the information becomes increasingly diffused, to the point of becoming a "whisper" above the 50% grey level. | 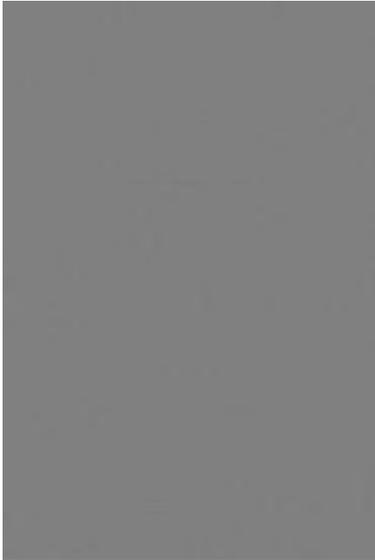 |
| Blur 13:<br><br>Parameters in pixel values:<br>spread = 2, diffusion = 3<br>file size = 262.0 Kbytes | 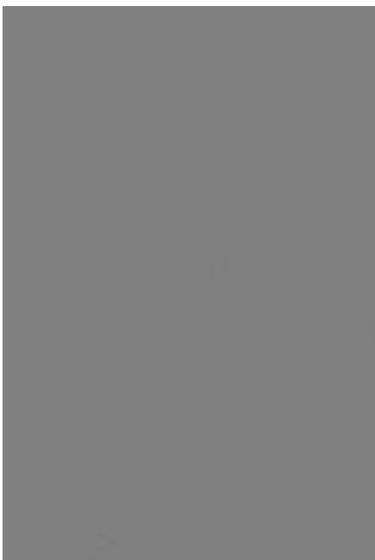 |
| Base:<br><br>file size = 1.49 Mbytes<br>JPEG quality 95% in GIMP | 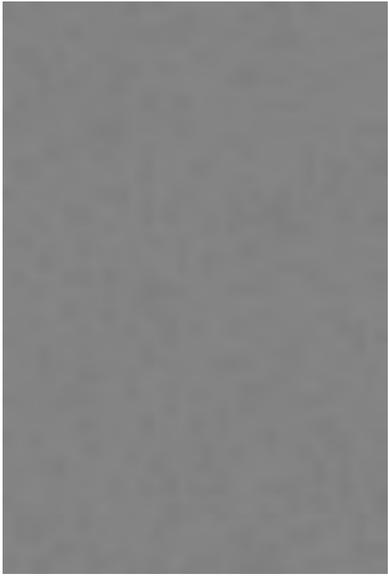 |

The original TIFF file size was 23.4 Mbytes, the compressed stack as in Table 1 is of total size 3.12 Mbytes. The compression ratio here is 7.5 times. The original image, the reconstructed image, and their difference is shown in Table 2 below.

**Table 2. Image comparisons for Case 1.**

| Original image | 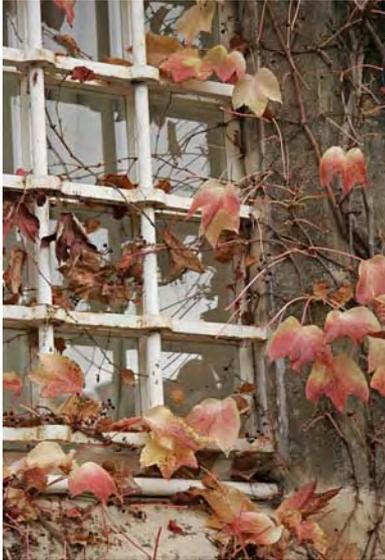 |
|---|---|
| Reconstructed image from the stack shown in Table 1 | 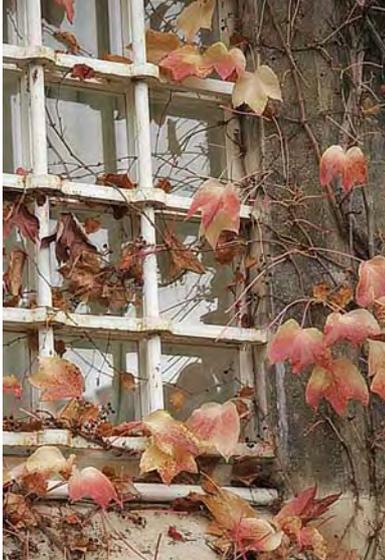 |

| Difference image | 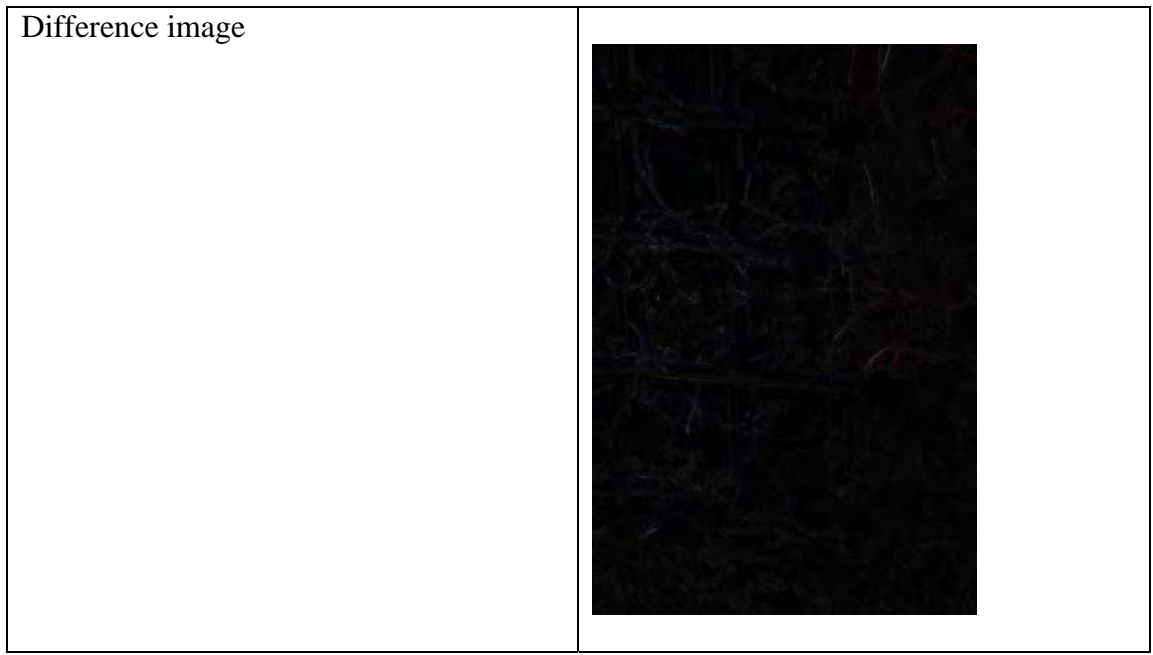 |
|---|---|

Upon close examination the difference image is not completely black, as expected, because the base image has been compressed using the JPEG algorithm which is lossy.

An idea to overcome this problem is to further store another smaller stack of Gaussian Blur decomposition working on the difference of the reconstructed image from the original TIFF image. The secondary stack can be small, and can be used to further add correction to the reconstructed image.

It is also noted that the reconstructed image is more equalised in contrast and exposure, which is the result of the blurring operations. It is also noted that there is some over-sharpening, with some halos appearing around prominent edges. This is the result of the compression of the final base image, which no matter how almost grey it is, it still retains enough information it appears to influence the final result.

Finally, the size of the intermediate blur levels was not reduced by as much as it is possible to achieve as this was the very first experiment conducted with the new algorithm in order to gain understanding of its operation. It would be possible to reduce more dramatically the size of the blurred JPEG files, as they are of no important significance in themselves individually, e.g. by up to 2-4 times, thus gaining overall compression ratios of 15 to 30 times. This is demonstrated in the next case study.

The reconstruction of the image can be done in any desirable order, top to bottom, bottom to top, or in a random fashion. Below, in Table 3, the reconstruction from bottom to top is shown, while Table 4 shows the reconstruction from top to bottom.

**Table 3. Reconstruction bottom to top (viewed left to right, top to bottom, i.e. row-wise).**

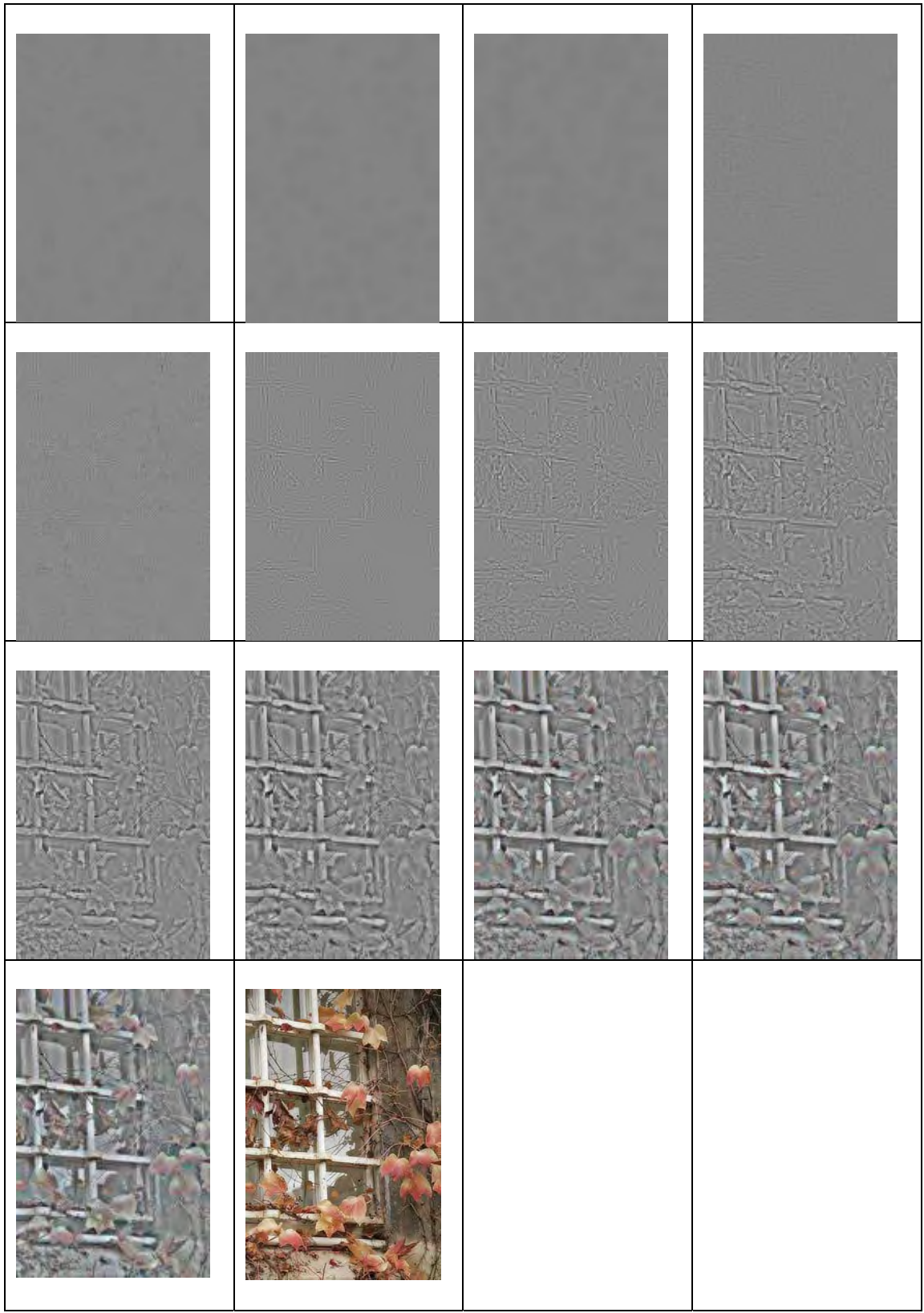

**Table 4. Reconstruction top to bottom (viewed left to right, top to bottom, i.e. row-wise).**

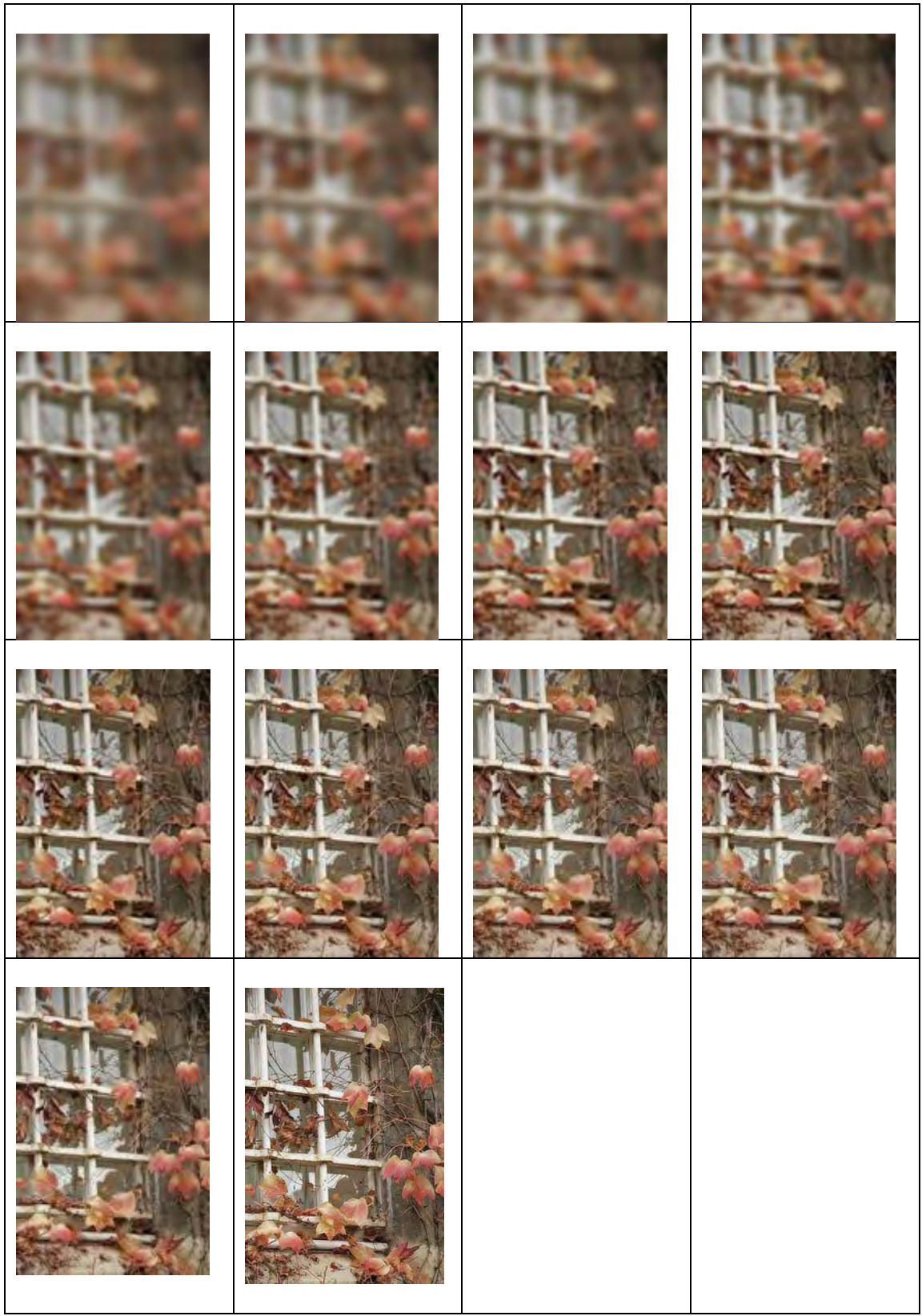

The top to bottom reconstruction starts creating the detail of the image and adds the colour finally to images that are focused almost all the way (also revealing information at each level). The top to bottom proceeds by unfocused images in full colour that are progressively becoming more focused. Observing the latter sequence as it unfolds one has the feeling of moving from outside into the image, i.e. it creates a stereoscopic perception of the image as it focuses.

**3.2 Decomposition of a noisy image**

This image was produced in 16-bit colour depth TIFF from scanning of a printed photograph. The TIFF was mapped to 8-bit colour depth as use of JPEG compressions restricts colour depth to 8-bit. Table 5 shows its decomposition.

**Table 5. Decomposition of a noisy image (viewed top to bottom, right to left, i.e. column-wise).**

| Blur 1: Diffusion 1000<br>File size: 15 Kbytes<br>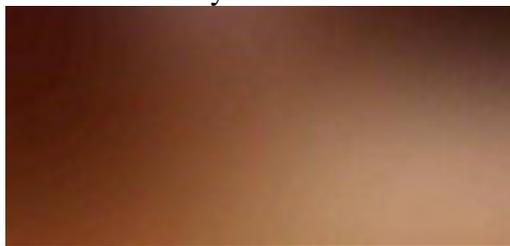 | Blur 8: Diffusion 16<br>File size: 15 Kbytes<br>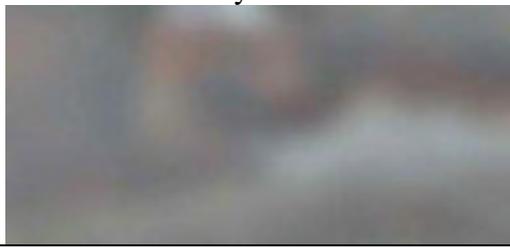 |
|---|---|
| Blur 2: Diffusion 500<br>File size: 14.8 Kbytes<br>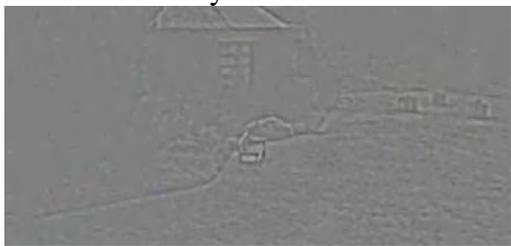 | Blur 9: Diffusion 8<br>File size: 15.6 Kbytes<br>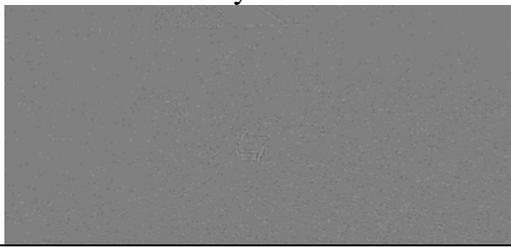 |
| Blur 3: Diffusion 250<br>File size: 14.9 Kbytes<br>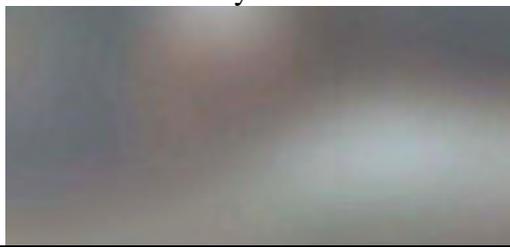 | Blur 10: Diffusion 4<br>File size: 15.7 Kbytes<br>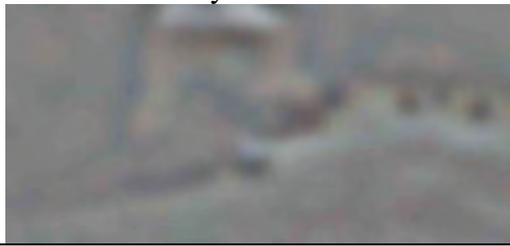 |
| Blur 5: Diffusion 125<br>File size: 15 Kbytes<br>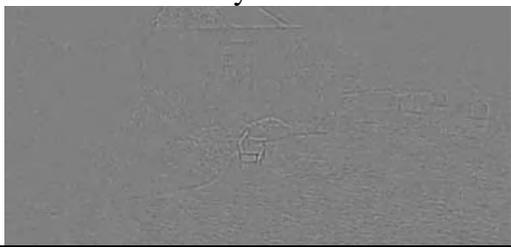 | Blur 11: Diffusion 2<br>File size: 17.1 Kbytes<br>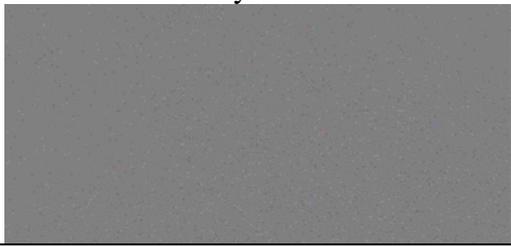 |

| Blur 6: Diffusion 60<br>File size: 14.9 Kbytes<br>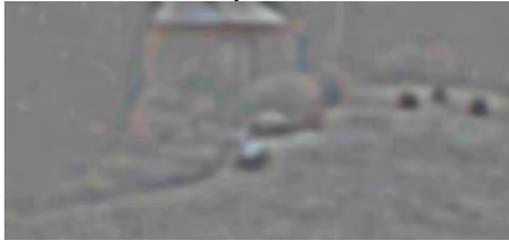 | Blur 12: Diffusion 1<br>File size: 16.4 Kbytes<br>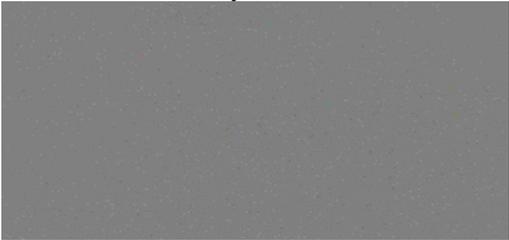 |
|---|---|
| Blur 7: Diffusion 30<br>File size: 15 Kbytes<br>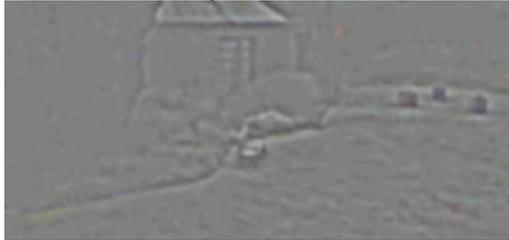 | Base: JPEG quality 50%<br>File size: 229 Kbytes<br>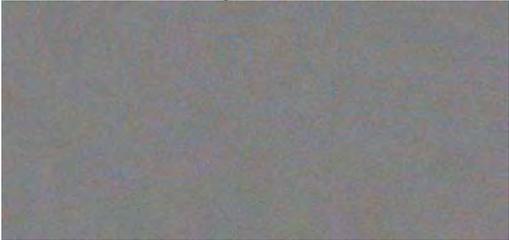 |

The original image had a size of 5.99 Mbytes in TIFF format, the total Gaussian Blur decomposition stack presented has a size of 0.399 Mbytes. The compression factor achieved is 15. The major obstacle in achieving better compression was the base image. As the image originally was very noisy (degraded negative film of 1000 ISO) a lot of noise has accumulated in the base image. Noise reduction techniques can be employed on the noisiest levels (from levels 7-8 downwards) so as to further assist in the compression of the base image, and to achieve far better compressions even in such noisy cases.

Some smoothing (further blurring of levels 9 through to 12, with mild parameters of 10 pixels diffusion, applied on the decomposed images) was used as an experimental approach to reduce the noise of the reconstructed image. The results along with the original, reconstructed, difference and noise reduced images are shown in Table 6 below.

Another observation about the nature of the noise accumulated in lower levels of the decomposition and the final base image, is that the noise is not in grey mode, but rather it is RGB (colour speckles). Perhaps turning the base into greyscale could also help in reducing the perceived noise in the reconstructed image.

**Table 6. Comparisons in the reconstruction of a noisy image.**

Original image

It is noted that the highlights of the image, predominantly at the roof tiles and the bench are "blown" (saturated), with also some specular highlights blown on the grass area

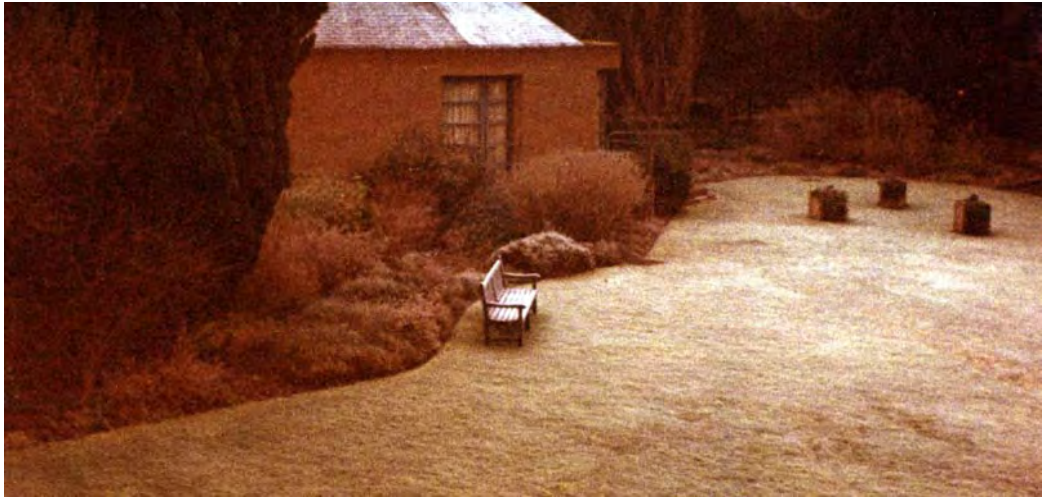

Reconstructed image

The reconstructed image is almost identical to the original, with equal amount of noise perceptually. It is notable though that a difference here is that the blown highlights have been "equalised" by assigning them the colour that is representative of nearby regions, namely the blown highlights in the tiles have turned reddish which was the approximate colour of the original scene. Of course this constitutes distortion of an original image, but then again, blown highlights are considered to be "dead" pixels with no information content in the first place.

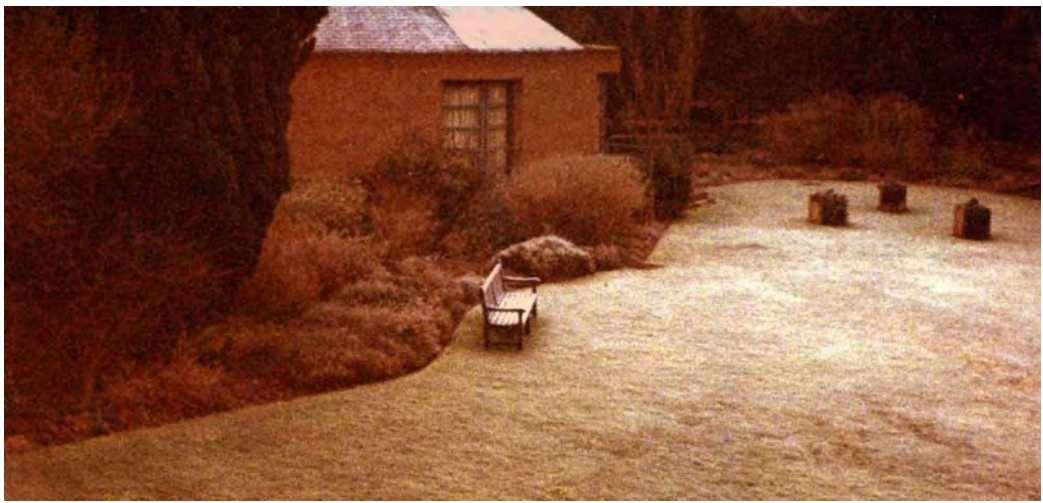

<u>Difference image</u>

The difference is exclusively located in the blown highlights region. The region contains some information in terms of contrast differences that constitute the edges of the tiles and the bench which are somewhat "washed out" in the reconstructed image. A finer tuning, in an iterative predictor-corrector mode, in the decomposition could preserve that detail if required.

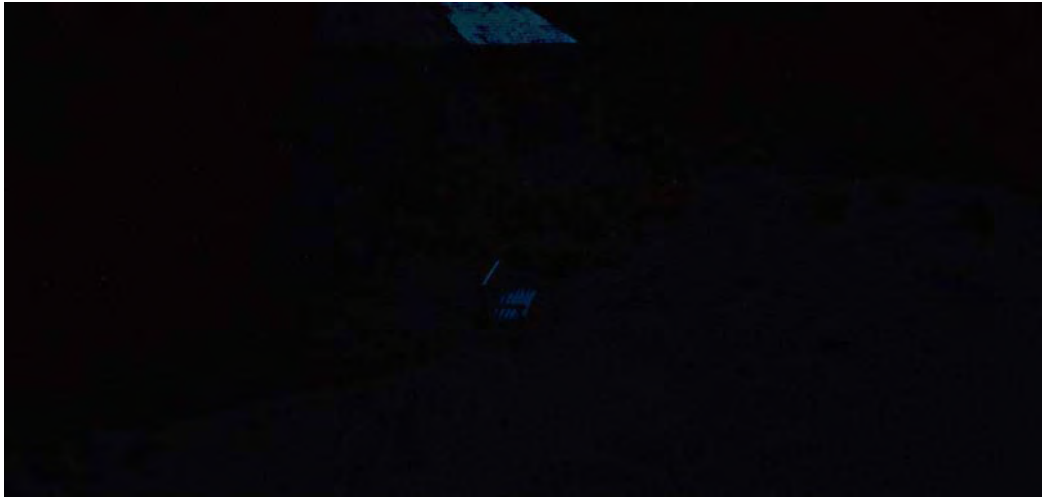

Noise reduced image

To reduce the noise in the reconstructed image, the layers numbered 9, 11, and 12 were *blurred* with a Gaussian diffusion coefficient of 10 pixels in the decomposition. Layer 10 was not blurred as it was found to contain vital edge information although it was also relatively noisy. Further to this, the base image was converted into greyscale to remove completely the RGB speckles observed there. It was then blurred with a Gaussian diffusion coefficient comparable to the observed noise, set at 3 pixels. The base from its original uncompressed TIFF stored separately, was compressed using JPEG at quality of 50% resulting in a file size of 101 Kbytes.

The overall enhanced image decomposition (accounting only the difference in the new grey base image compression difference from the originally reconstructed image) yields a total decomposition/compression size of 0.271 Mbytes. The corresponding compression factor from the original TIFF image is 22.1, without significant loss of quality due to the dominant noisy element in the original image.

To further enhance the image, it is possible to strengthen the contribution of the decomposition layers carrying edge information, after some more sophisticated noise reduction algorithm is applied to them (perhaps in greyscale variants of the layers). This will have to be done prior to compression of the blurred layer into lower quality JPEG format. In effect, the proposed procedure is a more sophisticated extension of the "Unsharp Mask" technique.

Where absolute quality of the image is paramount and no compression is of interest, then the blurred layers can be stored in raw format (uncompressed) so as to allow future specialised algorithms to focus on the image enhancement and analysis aspects offered by the Perceptron Algorithm.

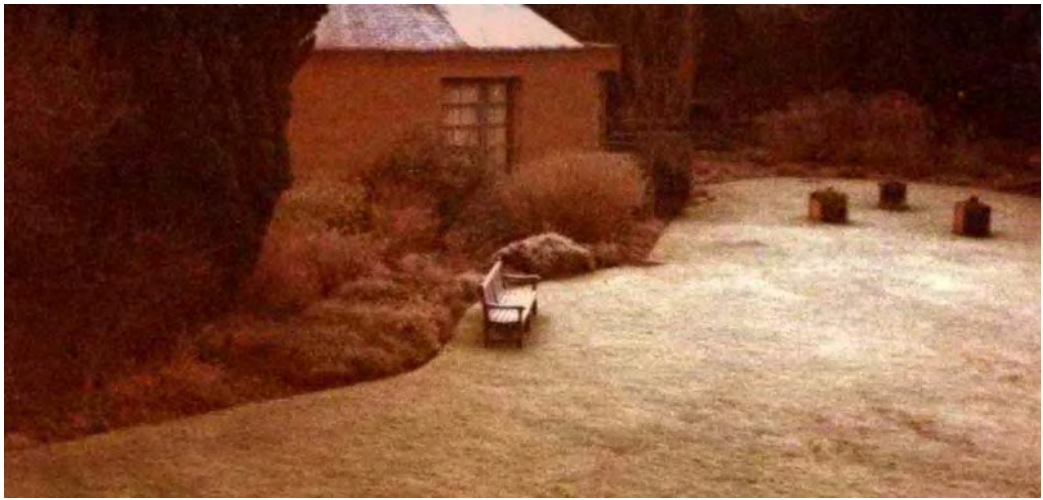

## 4. Multichannel image decomposition and related considerations

Most images of interest in the modern digital medium are composites, i.e. they result from the superposition of similarly captured images, either in the same window of frequency sampling, or in fact resulting from the superposition of different spectral windows' sampling, such as RGB images obtained with optical capture devices. The discussion is by no means exclusive to images, and can be generalised to any such form of signals. To highlight the issues of composite signals we focus in this section exclusively on RGB photographic images.

The first observation is related to the Perceptron Algorithm application in the 3 channels separately. As different densities of information are captured in the three channels, it stands to reason that a further improvement in the efficiency of the algorithm, both for compression and image analysis, would be to apply it separately to each channel thus treating the R-G-B channels as three separate images (which they are).

For digital cameras, the most favourite (and cheapest) way to construct a sensor is the Bayer array, where R, G, B, sensitive pixels are arrayed in a checkerboard fashion. There are twice as many green pixels as there are individually red or blue, such that each red or blue sensor pixel is surrounded by 4 green ones. The colour distribution to all pixels is conducted via some form of appropriate interpolation over the whole array thus yielding a full resolution in RGB over the whole sensor matrix which is in effect interpolated, except for the channel value where there is an actual R-G-B pixel sensor sampling the colour luminance in that location. It is possible to envisage that the Perceptron Algorithm can serve as a superior interpolator if the pixel values are not interpolated but create a sparse image for the R, G, and B channels separately.

The Gaussian Blurring operations will spread the values over the entire matrix if applied to each channel separately and then the three (potentially also much more efficient decompositions) are superimposed when the image is to be composed as a whole. This might also allow a much more significant compression of the image at the source (i.e. the camera, which with a little more processing capability will store a smoother image, with significantly reduced size in a "lossless" fashion on the memory card, whose capacity is used in the most efficient manner).

A final issue discussed in this section, concerns the RGB distribution of values for the Perceptron Algorithm. Some colour separation is observed as the layers are decomposed, and it was mentioned that the very first layer seems to take away all of the colour from the image, yielding mostly grey images in later layers. We turn our attention here on the final base images as produced by the algorithm. Table 7 below shows some data collected from a few decompositions.

**Table 6. RGB analysis information for base images obtained with the Perceptron Algorithm**

| Base | Mixed image RGB information with high blur (1000 diffusion or more) | RGB tonality distributions |
|------|-------------------------------------------------------------------|----------------------------|

| **Noisy Image**<br><br>Image source is an EPSON PRO 35mm/platen scanner. | Pixel values / RGB<br>Red: 128 / Red: 50 %<br>Green: 128 / Green: 50 %<br>Blue: 128 / Blue: 50 %<br>Alpha: 255 / Alpha: 100 %<br>Hex: 808080<br><br>The blended image average is indeed a 50% grey image, as shown in the table above.<br><br>However, the histograms to the right indicate that the spread of effectively what is noise in the base image is similar for the G and B channels, but it is more spread for the R channel.<br><br>In a way this shows that noise although mostly attributed to the B channel, in some cases it can manifest itself in any of the other ones.<br><br>Also noise in the overall image is shown in the overall value (luminance) channel on the top in the right. The spread is centred at an average value which is shifted slightly to the highlights. This indicates that application of the Perceptron Algorithm will make the base not go more grey, but as one new blur is subtracted from the working base it will shift the noise in other locations. Clearly such cases call for a dedicated application of a noise reduction algorithm at each working base layer as it is detected to be noisy.<br><br>Furthermore, an edge detection algorithm can be applied to a base image to filter out possible edges. Noise reduction methods applied to the edges' image can remove noise and the greyness can be restored by suitable re-composition with the originally noisy base image. This may be applied at other layers of the decomposition, obviously, to enhance both clarity and allow futher deeper compression without loss. | 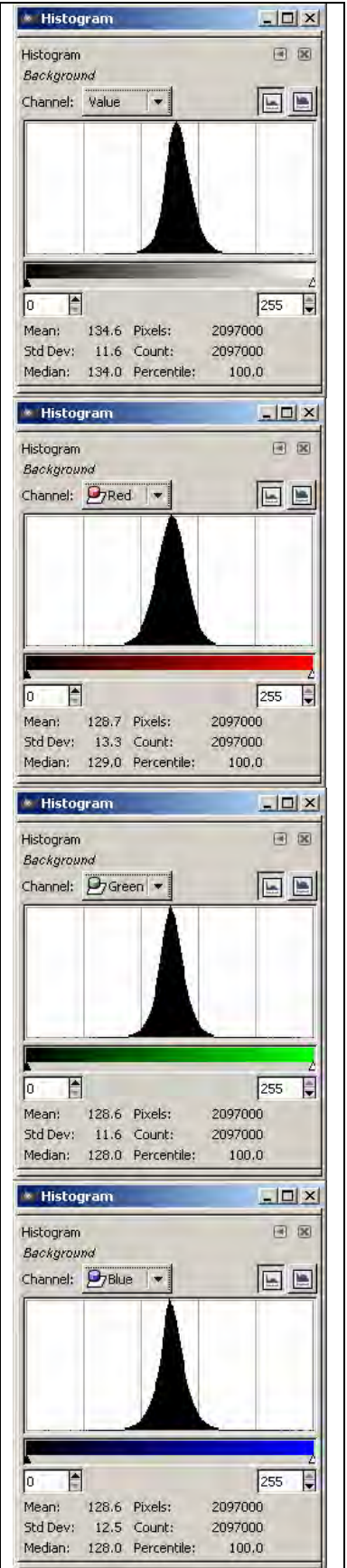 |

| **Low noise image**<br><br>Image source is a CANON EOS20D 8.2 Mpixel dSLR camera. | 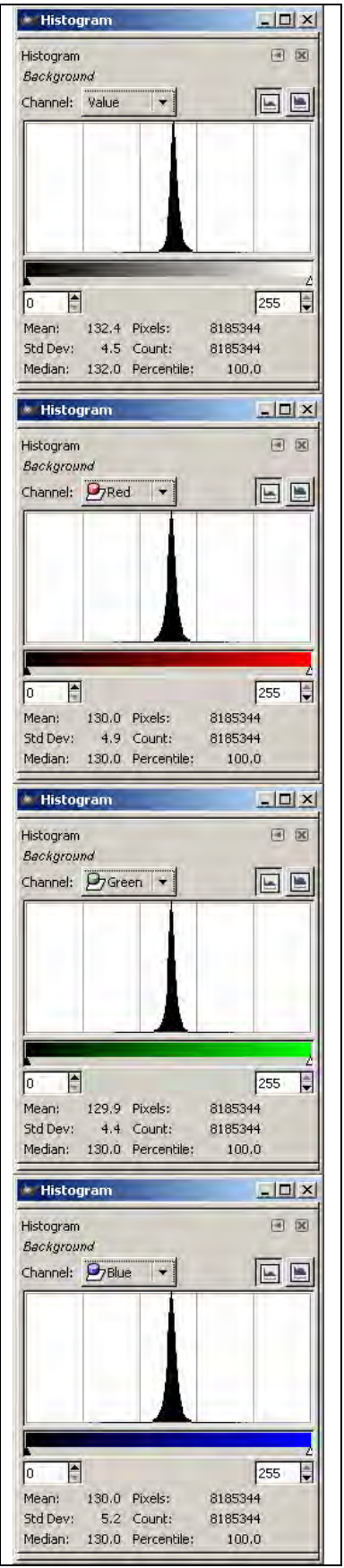  Here the overall luminance distribution is similarly shifted to the highlights as the RGB individual channels.<br><br>There is inevitably some noise in the image, but the shift in the base image distributions is because there exist still edges' information in the image.<br><br>Interestingly, there is slightly more spread again in the R (red) channel. This leads one to consider whether the hypothesis for considering the B (blue) channel as the noisiest is correct. | |

# 6. Distance and size computations based on focal and aperture blurring

When taking a photograph, or any other form of image, a focusing apparatus is used to obtain a sharply focused object onto the sensor (film, recording media, even in other forms of signals such as sound).

In photography this is achieved by the combination of the lens and its aperture opening which is of varying diameter. The size of the object obtained is a function of the focal length used and the distance of the actual object from the lens.

The blurring of objects in front and behind the object where we focus on depends on the aperture diameter/opening (iris) and the lens combination. This blurring is in effect out of focus objects and the effect is akin to Gaussian diffusion. Diagrammatically, shown as unequal cones of focusing (in photography this is related to the "hyperfocals distance" and related concepts and phenomena) is the scene in front of the object we focus on, the object itself and the scene behind it. This is shown below in Figure 1 below.

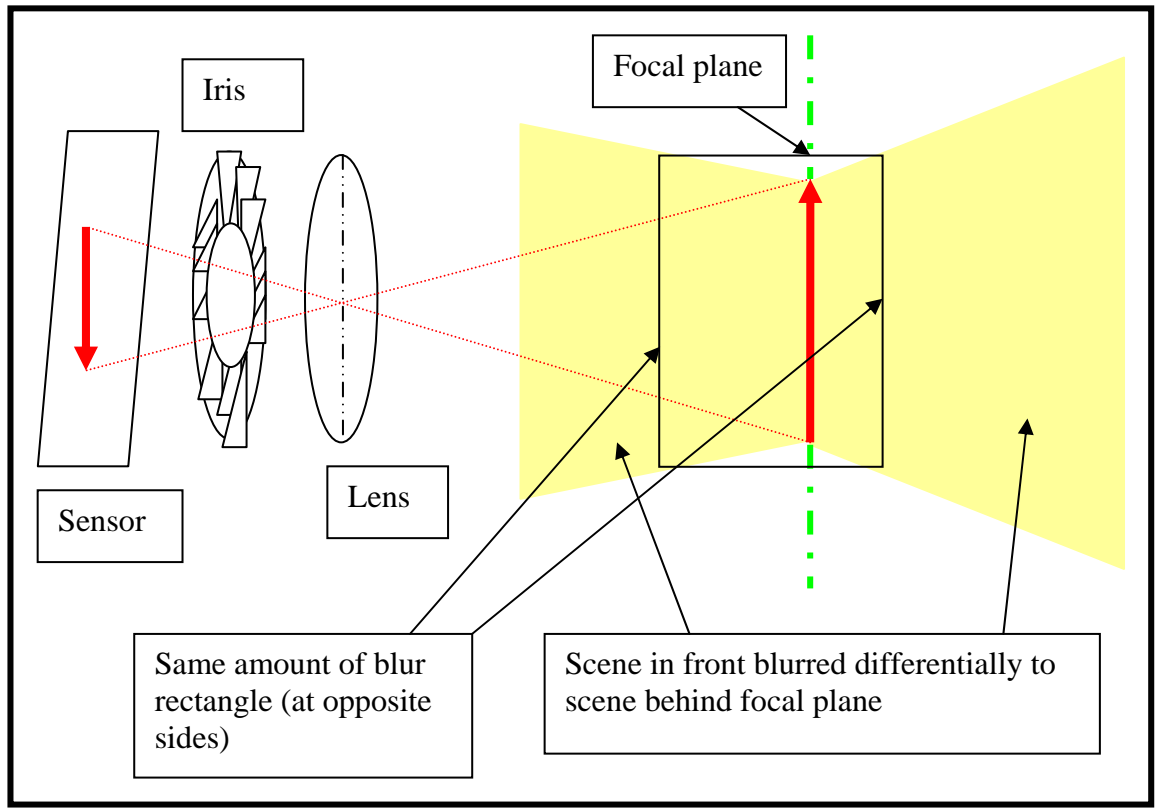

**Figure 1.** Depth and size estimations using blurring (out of focus regions)

It is proposed here that using the Perceptron Algorithm (incremental Gaussian Blurring) to produce depth information and a sense of stereoscopic image reconstruction; this can be coupled with the orientation of shadows manifested as contrast regions identified in the decomposition (mention elsewhere in this paper).

To do this, we need to combine the following information:

    (a) Focal distance

    (b) Aperture size
    (c) Recorded object size
    (d) Model the scene blurring: in front and behind the focal plane
    (e) Apply the Perceptron Algorithm, observing that the most persistent edges remaining at the base image must be coming from the *most focused object*, as its edges will be the most sharply defined in the recorded image (or signal, if equivalent principles can be found for any type of signal analysis).

A combination with the orientation of shadows in relation to the blurring might be able to place correctly parts of the image as in front or behind the focal plane, thus producing a stereoscopic analysis of what is perceived as a flat image to begin with. No illumination will be perfectly perpendicular, and reflected light will create shadows inevitably, thus it might be possible to exploit such properties to reconstruct depth and size information.

The working principle is that blurred parts of an original image will blur faster when Gaussian blurring is applied and hence will be subtracted at earlier stages of the Perceptron Algorithm. Edge analysis of a top down recomposed image at each stage may be able to identify which edges are further from the focal plane. The amount of Gaussian blurring (diffusion coefficient) could be used as a measure of distance from the focal plane, along with the other parameters of the imaging system mentioned in the list above.

To obtain more information and be able to relate differentially what is in front and what is behind, one may also consider taking photos of the same scene with mutliple focal points (focusing at different depths and also combinining these with different aperture openings per focal point, one for each shot), and comparing the decomposition of the images together.

Motion also causes blurring such that moving objects will appear blurred in a specific way in the recorded image. Motion analysis may combine other forms of blurring, such as "motion blurring" and "radial blurring" found in most standard image manipulation packages, to design algorithms that analyse properties of moving captured objects (e.g. estimate velocity, direction, recompose their shape by moving pixels "backwards", etc.).

Finally, with the above observation, other forms of blurring applied to still images either alone or in tandem with each other, could reveal other aspects of information imbedded in an image. For example, if motion blurring is generalised to include a spherical angle orientation of the motion along with the perspective distortion when this motion is recorded on a two-dimensional medium, one could infer the spherical angle direction along which a recorded object was moving (in or out of the page, along with the angle of rotation on the plane of the recording medium).

For example, if the motion is completely in the opposite direction of the spherical motion blurring analysis applied (assuming motion in a straight line for the duration of the exposure), then the object will be "recomposed" bringing the pixels back to form sharp edges. Then, if on top of this Gaussian blurring analysis is applied it might be possible to reconstruct information about its shape, size, and distance from the observation point (camera location for example).

In a further extension, one might consider real-time tracking systems that to enhance their ability to correct and perceive motion utilise a combination of spherical angle blurring analysis along with Gaussian blurring. If two moving imaging/sensor systems are placed close to each other then their difference in distance from each other can create a stereoscopic depth perception along with analysis of motion and tracking. This might be the combination of "hardware" and "software" that makes the eyes (with the ability to vary rapidly, at will, the iris aperture opening and focal length of the lens) and the brain able to analyse such information rapidly.

Such discussion is only presented here in broad terms, hoping it could form the kernel for future research in the field.

## 6. Summary of observations and future ideas for research and development

The ideas specifically are outlined in summarised form as:

1. Digital image compression without loss, or with tunable loss depending on the precision of the reproduction device targeted. The image is decomposed in a mathematical-algorithmic way, which is computationally cheap and simple, into a multilayered array (either a stack or a tree structure) of simpler images which can be compressed to the order of 10 Kbytes each, indicatively. Tunable loss can be achieved by setting a threshold tolerance for the difference of the reconstructed image to that of the original, so that the algorithm may decelerate or accelerate the reduction in the blur coefficient so as to produce fewer or more layers in the decomposition to match the desired accuracy of reconstruction.

   This idea can lead to an extension of the JPEG compression algorithm to become multilayered to accommodate the Gaussian Blurring stack; it would also be interesting to see an extension of the JPEG protocol to accommodate 16-bit colour depth. If the derived image by any JPEG compression uncompresses proportionately to the file size, then since the operations of the reconstruction in the Perceptron Algorithm are mere additions effectively, then since also the resulting size of the stack is smaller than 100% JPEG quality, it stands to reason that the new compression proposed would be faster in reconstructing compressed images (and in general other types of compressed signals) than that top quality JPEG compression reconstruction (and potentially with much smaller losses in quality).

2. Demonstrations are available where manually an image was compressed from TIFF format of 23Mb (8-bit colour quantisation, 8.2 Mpixels resolution) to approximately 30 times smaller size using the standardised image manipulation software GIMP, produced by the GNU organisation.

3. Noise suppression and compensation properties of the array of images, either in transmission, storage, or when restoring older images. The decomposition-compression method assigns automatically most of the noise of the original image into readily identifiable sub-layers of the decomposed image, and in so doing it becomes very easy to apply noise correction algorithms only to the layers that are automatically identified as noisy. Since there is an additive property in the reconstruction of the image from the layered stack or tree-structure, then the noise-reduced or corrected image will not suffer blurring or loss of edge detail, to the extent that ordinary methods will do when applied to the entire original image.

For transmission and storage errors, since the image is decomposed into a stack of layers, if an error would occur at a random location in some layer(s) the correction would come from the additive reconstruction by the superposition of the remaining layers. In other words since the total pixel luminance and hue values come from the superposition of many pixels merged into a single pixel, error in a few of them can be compensated in this fashion quite effectively. Further to this, due to the compression capability of the algorithm, if an image is reduced in storage size significantly it can be afforded to re-transmit it more than once depending on the criticality of the application (e.g. transmission from satellites, space probes, etc.).

4. Exploiting the smoothing of the first layers into highly blurred images and of the later layers where mostly edges are predominant, it might be possible to *enlarge* smaller images (in the number of total pixels available) and obtain a similar quality higher resolution image without jagged structures or artefacts appearing in them. This is likely to be so, because the blurred top layers carry little structural information so a smooth (e.g. cubic) interpolation might be very good for them. Similarly for "edgy" layers, in the later stages, which are also almost 50% grey or noisy in other areas, the edges can be enlarged relatively accurately and so the overall enlarged and recomposed image might be much smoother than other methods of enlargement working directly on the original image. In other words, a tunable layer-by-layer approach can produce high quality enlargements. Of course no new information can be generated from a smaller image, but it is expected that for reproduction purposes this method will perform very well. Coupled with noise reduction properties, it is possible to restore and re-examine older images and films.

5. Structural analysis properties of images by either examination of decomposed layers of the stack, or by partial re-composition of the image. Applications can be derived readily for Magnetic Resonance Imaging and CAT-scan images (medical and general analytical imaging in general) which (a) can be compressed dramatically to facilitate transmission and storage of medical records, (b) can be analysed automatically to identify structural problems for example in many medical diagnostic situations (see below: pattern recognition applications).

6. Exploitation of super-compressed images to reduce storage needs by orders of magnitude, and to reduce transmission speeds over wireless or cabled transmissions with applications in:

    a. Internet decongestion by orders of magnitude in the transmission of digital images and scanned documents.
    b. DVD & film storage compression by orders of magnitude without further improvement or demand on storage media.
    c. Pay-per-view applications where entire movies in the new compressed format can be so compressed so as to facilitate rapid and cheap transmission to a home basis delivery of films (movies) by cable companies.
    d. Compression of images at the source, for example photographic, scanning, and imaging devices (cameras, scanners, medical imaging

systems, scientific imaging and not only restricted of course to visible light imaging systems).

7. Parallelisation schemes for the acceleration of the compression

    e. Splitting of the image in parts, with overlapping or non-overlapping segments (which can be automatically identified, depending on the needs and seriousness of the application) so as to assign each sub-image to a separate processor.
    f. Adjustable, predictor-corrector mode of operation where portions of high detail require more analysis which is recognised automatically, while others of smaller detail are analysed and compressed more rapidly.

8. Parallelisation schemes for the acceleration of the reconstruction

    g. In reverse order of items 5a & 5b above, the reconstruction can be produced by a parallel computing system which exploits the decomposition of an image in subparts.

9. Face recognition and scene recognition (pattern recognition in general), and accelerated database searching. The decomposition of an image in "layers" of varying complexity, each of which is much smaller and simpler when compared to the original image, offers a unique and very novel way for pattern recognition (with applications in security cameras, streaming real-time video, computer/robotic vision applications, motion analysis, etc.). The database searching scheme presented is based on the following steps:

    h. Since the decomposition can be standardised to produce in the same way a compressed decomposition of an image, these decompositions can be searched *horizontally (at the same level of decomposition layer)* with respect to a reference photograph/image decomposed in the same fashion. The comparison can be simple, such as simple subtraction of the pixel values of the corresponding "layer" of stored images with the corresponding "layer" of the reference image. A statistical measure of similarity can be thus derived readily, and potential matches can be selected out of the original larger set stored in the database.
    i. Once one tier of the comparison for the entire contents of the database is conducted, the comparison will collect the likely hits and analyse those for the next layer, doing so recursively until the number of matches is narrowed down to a few, or to none.
    j. The search can be done in parallel, for example by splitting the contents of the database and assigning them to different processors while giving to all of them the same reference (decomposed and compressed) image to search for. A coordination step is conducted at the conclusion of all the processors tasks for the same depth they have been assigned.
    k. It is noted that because of the way the Perceptron Algorithm is conducted slight variation in angle of presentation, colour and fine detail, will not affect much the pattern recognition capabilities of the algorithm. Where more detail will be invariably compared and searched for, minor corrections for skewness, rotation, etc. can be

applied to overcome these problems that can been serious issues in pattern matching.

The idea is demonstrated in Figure 2 below.

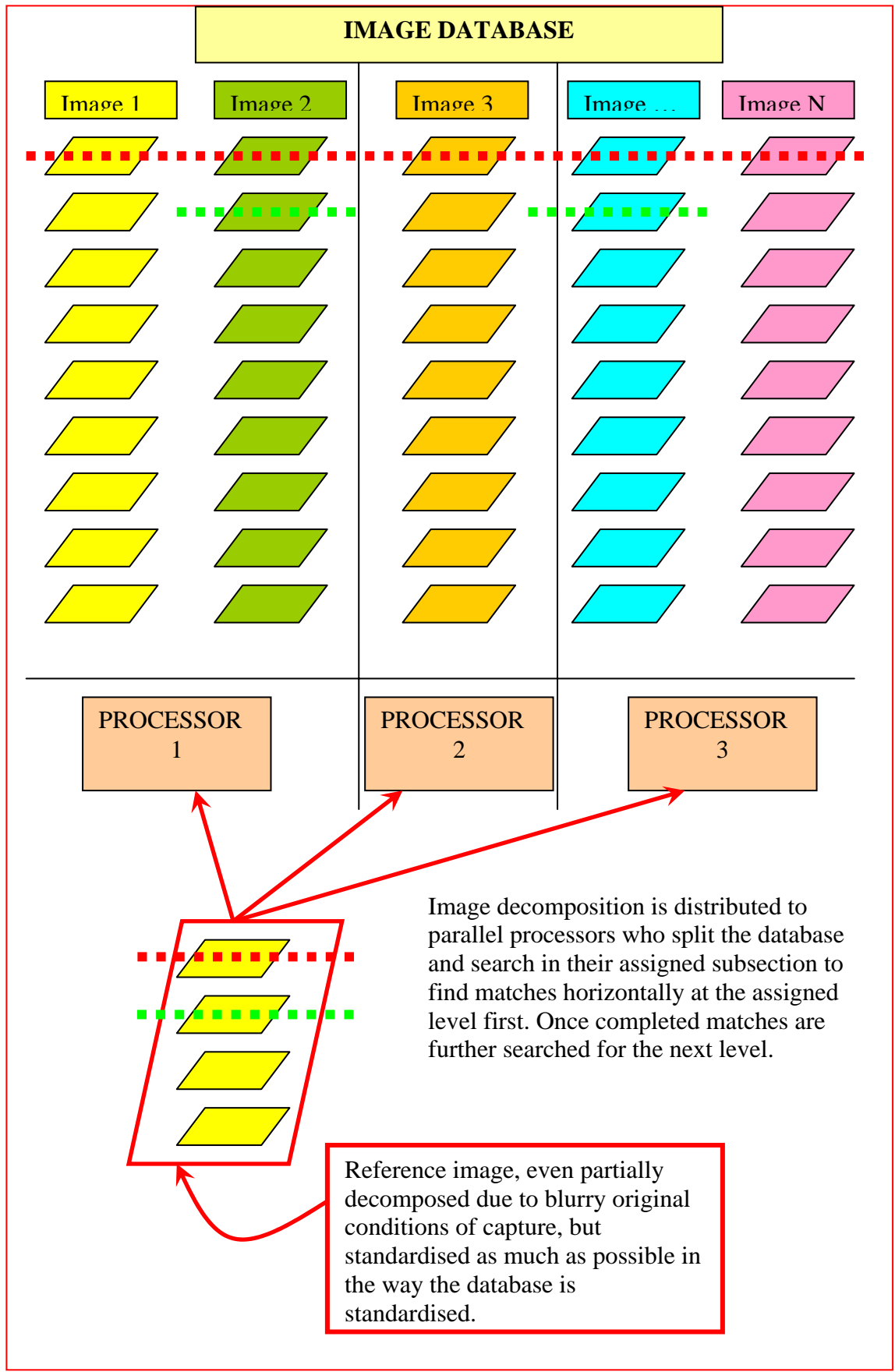

**Figure 2.** Database searching for pattern recognition schematic operation.

10. Medical applications and neural computing. Ideas regarding this are also being considered for future work, indicating here that the motivation for the Perceptron Algorithm was derived from observations and conjectures regarding how memory seems to operate in the human brain (as mentioned in the Introduction section). From this the algorithm presented was derived, and based on its successful demonstration in this work, it is further conjectured that novel advancements can be researched regarding both human neurophysiology as well as the design of new analogue or digital computing devices and algorithms motivated by new insights. In a way, the successful few first applications of our algorithm may serve as verification of the ideas on neural structures (e.g. the hippocambus) for memory and processing in the biological brain.

11. Depth analysis information. The information obtained by the decomposition of the Perceptron Algorithm is possible to be related to contrast differences at various levels of blurring subtractions. If this information can be related to the relative placement of objects *in depth, in relation to each other,* then it is possible to conjecture that appropriate research into the methodology can be used to reconstruct and enhance stereoscopic information obtained from two-dimensional images alone. Already, upon close examination of the reconstructed images as they evolve and the blur levels, creates a feeling of relative depth of the objects perceived in the field of view.

## 6. Conclusions

The purpose of this paper is to serve as an initial exposure to the ideas underlying a novel and very simple algorithm for image compression and analysis. The key concept is the use of iterative Gaussian Blurring and subtraction of the blurred image from the working image until a nearly 50% grey image is obtained as a base. Compared to other methods that may exist, Gaussian Blurring is a very fast, very simple, and efficient technique which has been shown to yield promising results in this work.

It is believed that the ideas outlined here can have immediate profitable results by the application of the algorithm in the fields identified above, to begin with in image and signal compression.

Further to these, the new approach is believed that it can create a chain-reaction of innovations in many fields, without causing technological pressure for more advanced equipment, although it is also believed that it will lead to such innovations as well.

As a final example one might consider that since the algorithm can restore imperfections in an image, within reason, it could reduce significantly the cost of construction of digital camera sensor arrays (not only restricted to optical sensors of course) by relaxing the tight homogeneity in their sensors, perhaps even producing acceptable enlargements by lower number of pixels, and even by not so homogeneously arrayed arrangements of sensor pixels.

After all, the eye itself has a finite number of rods and cones with "imperfections" in their placement, and yet when we view scenes we do not perceive jagged edges or homogeneity defects. It is the conjecture of the author that the correction in perception is "software" related. Perhaps a step towards understanding these issues might be the work presented in this paper, along with the ability to recognise and store in memory thousands of different shapes and patterns, and to discern them even in blurry conditions.

## References

The work here is primarily experimental and original, stemming from work in retouching photographs in GIMP and Adobe Photoshop$^{TM}$, primarily extending the standard technique of "Unsharp Masks" in an iterative manner as used in sharpening and "Local Contrast Enhancement" of images.

In brief bibliographical searches no similar approaches with such a wide breadth of recognisable possible applications and connections to both software and technology has been found, and to do so would have produced an enormous amount of citations to similar and parallel approaches, which in our opinion have nonetheless never been put together in the way attempted in this work; this by no means does not recognise the work of numerous researchers in the field. Future research stemming from this paper will provide links with other techniques in the field of image, audio, and signal processing.

Hence no references are used in this work which represents mostly an experimental application of well-established and simple tools, available in most off-the-shelf image manipulation and processing software.